%% file: main.tex
\documentclass[10pt,twocolumn,letterpaper]{article}

\usepackage{cvpr} 

\usepackage{times}
\usepackage{epsfig}
\usepackage{graphicx}
\usepackage{amsmath}
\usepackage{amssymb}
\usepackage{tabularray}
\usepackage{multirow}
\usepackage{pifont}

\usepackage[pagebackref,breaklinks=true,colorlinks,bookmarks=false]{hyperref}

\RequirePackage[capitalise]{cleveref} 

\newcommand{\approach}[0]{RG-Gait }

\def\maketitlesupplementary
   {
   \newpage
       \twocolumn[
        {
        \centering
        \Large
        \textbf{Mind the Gap: Bridging Occlusion in Gait Recognition via Residual Gap Correction}\\
        }
        {
        \large
        \centering
        \vspace{0.5em}Supplementary Material \\
        \vspace{1.0em}
        }
        
       ] 
   }

\begin{document}

\title{Mind the Gap: Bridging Occlusion in Gait Recognition via Residual Gap Correction}

\author{
    \begin{tabular}{c}
        Ayush Gupta\hspace{3em} Siyuan Huang\hspace{3em} Rama Chellappa \\
        {\tt\small agupt120@jhu.edu}\hspace{2em} 
        {\tt\small shuan124@jhu.edu}\hspace{2em}
        {\tt\small rchella4@jhu.edu} \\
        Johns Hopkins University \\
    3400 N. Charles St
    \end{tabular}
}

\maketitle
\thispagestyle{empty}

\input{sec/abstract}

\input{sec/sec1_intro}

\input{sec/sec2_related_work}

\input{sec/sec3_method}

\input{sec/sec4_experiments}

\input{sec/sec5_conclusion}

\input{sec/sec6_acknowledgement}

\input{sec/supp}

{\small
\bibliographystyle{ieee}
\bibliography{egbib}
}

\end{document}

%% file: sec/abstract.tex
\begin{abstract}
Gait is becoming popular as a method of person re-identification because of its ability to identify people at a distance. However, most current works in gait recognition do not address the practical problem of occlusions. 
Among those which do, some require paired tuples of occluded and holistic sequences, which are impractical to collect in the real world. 
Further, these approaches work on occlusions but fail to retain performance on holistic inputs.
To address these challenges, we propose \textbf{RG-Gait}, a method for residual correction for occluded gait recognition with holistic retention.
We model the problem as a residual learning task, conceptualizing the occluded gait signature as a residual deviation from the holistic gait representation. 
Our proposed network adaptively integrates the learned residual, significantly improving performance on occluded gait sequences without compromising the holistic recognition accuracy.
We evaluate our approach on the challenging Gait3D, GREW and BRIAR datasets and show that learning the residual can be an effective technique to tackle occluded gait recognition with holistic retention. We release our code publicly at \href{https://github.com/Ayush-00/rg-gait}{https://github.com/Ayush-00/rg-gait}.
\end{abstract}

%% file: sec/sec1_intro.tex
\section{Introduction}
Gait is an important biometric feature due to its ability to uniquely identify individuals based on how they walk. 
Unlike other biometric modalities such as face or fingerprint, gait analysis can be performed from a distance and does not require physical contact, making it highly useful in various real-world applications such as surveillance, security, and healthcare. 
Several methods have been developed to extract identifying features from gait videos \cite{deepgaitv2, opengait, deep-gait-survey}. 
In practical scenarios, occlusions pose a significant challenge for these gait recognition systems, especially when the occlusion is extensive or occurs in key areas of the body.

\begin{figure}
    \centering
    \includegraphics[width=\linewidth]{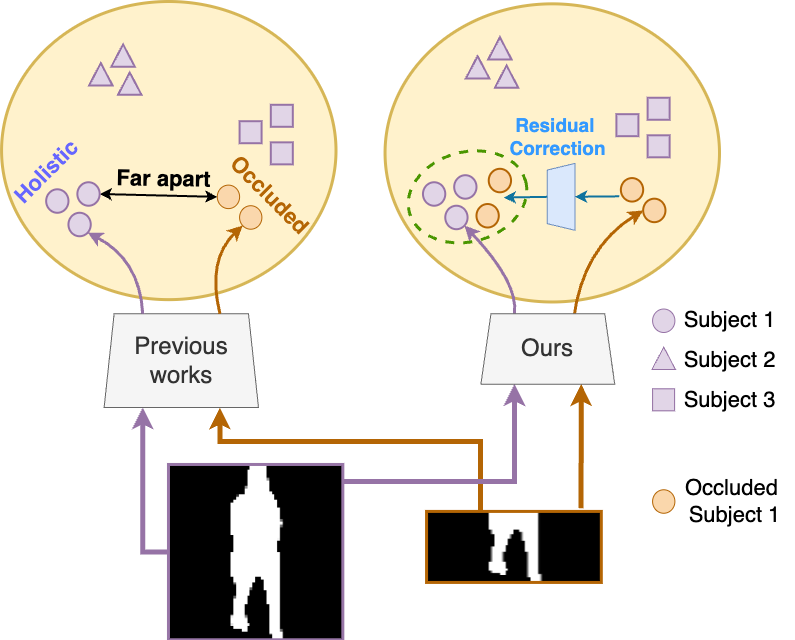}
    \caption{
    Illustration of the holistic retention problem.
    Optimizing the latent space for both occluded and holistic sequences with the same network proves to be challenging.  
    Thus, current works optimized for occlusion do not retain performance on holistic data. 
    To overcome this, we model the problem as a residual learning task, where the occluded feature is a residual deviation from the holistic features. 
    A separate network learns this deviation and refines the occluded features.
    }
    \label{fig:motivation}
\end{figure}

Recognizing these issues, a line of work exists in current literature specifically focused on the task of occluded gait recognition \cite{occ_aware, mimicgait}.
While these methods work well under occlusion scenarios, the occlusion-focused training of these models does not guarantee performance retention on holistic data. 
Training on occluded data can be seen as training on data from a different distribution than holistic data, and forgetting of source domain knowledge after finetuning on the target is a well-known problem \cite{forgetting}. 
We empirically confirm this forgetting problem and observe that current algorithms excel under occlusions but fail to maintain performance when the full gait sequence is available.
Further, some previous works also require paired data in the form of (occluded, complete) tuples to train their occlusion models~\cite{mimicgait}. 
We recognize that collecting such paired data is extremely difficult in the real world, further highlighting the need for methods that can handle occlusions without relying on paired supervision while retaining holistic performance.

To address this gap, we propose \textbf{RG-Gait}: Residual Correction for Occluded Gait Recognition with Holistic Retention. 
Our method introduces the idea of residual feature correction to gait recognition, where a correction feature improves the gait signature under occlusions while retaining holistic performance.
Specifically, we model the task as a residual learning problem, where the occluded gait features are one residual deviation from the (well-clustered) holistic features.
We employ an additional network which learns this residual and subsequently integrates it with the gait representation. 
This ensures that the features remain discriminative even when key parts of the body are occluded. 

One key aspect of our approach is the use of an occlusion evaluation module (OEM) to regulate the learned correction feature. The OEM, inspired from previous works \cite{mimicgait}, identifies the occlusions present in the input data.
The residual is adaptively integrated into the final gait signature based on the occlusions detected by the OEM.
This adaptive integration ensures that the correction feature has minimal effect on the gait signature in the absence of occlusions, thus preserving performance on holistic data.
As a result, \approach can adaptively improve recognition under occlusions without sacrificing recognition accuracy on complete data, offering a robust solution for real-world gait recognition.

Our method does not require paired (occluded, complete) data, making it more practical for real scenarios.
Further, our approach is model-agnostic, so it can be used with any gait recognition backbone to boost its performance on occlusions while enforcing holistic retention.

In summary, our main contributions are as follows:

\begin{itemize}
    \item We propose \textit{RG-Gait}, a novel model-agnostic method for occluded gait recognition which models holistic features as a residual deviation from occluded features. 
    \item We propose an \textit{adaptive feature integration} method to fuse these residuals in the gait signature, ensuring holistic performance retention without paired occlusion supervision.
    \item We evaluate our \textit{model-agnostic} approach on the challenging GREW, Gait3D and BRIAR datasets with multiple backbones, and achieve state of the art performance on occluded gait recognition while preserving holistic performance.
\end{itemize}

%% file: sec/sec2_related_work.tex
\section{Related Work}

\subsection{Gait Recognition}
A person's walking style, or gait, is an important indicator of the person.
Gait has been measured using wearable kinematic sensors~\cite{gait-unique, gait-sensor-2, gait-sesnor-1, gait-sensor-2-ramneet}, which become important in the medical domain for diagnosing diseases~\cite{diagnose-1, diagnose-2, diagnose-3}.
Gait has also been shown to be unique~\cite{gait-unique} to a human like a fingerprint, hinting at the viability of gait being used as a non-invasive method of person identification from a distance. 
This requires purely vision-based gait recognition, for which several algorithms have been proposed~\cite{deep-gait-survey, opengait, deepgaitv2, guo2023multi, vmgait, gaitcontour}. 
The introduction of many large scale datasets~\cite{grew, gait3d, casiae, briar, FVG-dataset} has helped in significant advancement of these algorithms. Indoor datasets collected in controlled environments~\cite{casiae} propelled the initial advancements, and now several datasets focus on outdoor or unconstrained gait recognition tasks~\cite{briar, grew, FVG-dataset, gait3d}. Occlusion is an important problem in these unconstrained scenarios; however, these datasets and algorithms do not specifically target this problem. 

\subsection{Occluded Gait Recognition}
Recognizing the occlusion problem in outdoor or unconstrained scenarios, some works~\cite{singh2022hybrid, Xu_2023_ICCV, xu_occlusion-aware_2023, uddin2019spatio, occ_aware, mimicgait, occ-reid1, occ-reid2, occ-reid3, occ-reid4} explicitly target the problem.
Some of these focus on the reconstruction of the occluded silhouette using generative techniques~\cite{singh2022hybrid, uddin2019spatio}. Some other works~\cite{xu_occlusion-aware_2023} focus on estimating an SMPL 3D mesh model of the human body, through which the missing body parts can be inferred. Most of these approaches simulate occlusions on indoor datasets and are not easily extendable to outdoor, long-range and turbulent data since they are sensitive to the reconstruction techniques.
Because of the difficulty in collecting real occluded data, some works simulate occlusions on outdoor datasets~\cite{Xu_2023_ICCV, occ_aware, mimicgait}.~\cite{Xu_2023_ICCV} perform silhouette registration to preprocess the input, but work with limited types of occlusions. 
Some works~\cite{mimicgait} require paired occlusion-holistic data for training their algorithms.
While these approaches work well with different types of occlusions, they are not able to maintain performance on holistic inputs.
However, our approach is designed for occluded gait recognition while retaining holistic performance on potentially long-range and turbulent data.

\subsection{Residual Correction}





Residual learning, as proposed in ResNet~\cite{resnet}, introduces skip or residual connections within layers of a neural network, allowing it the flexibility to either compute deeper, more abstract representations or directly propagate the identity function when appropriate. 
Residual Transfer Networks~\cite{residual-transfer} extend this concept specifically for domain adaptation, hypothesizing that the discrepancy between source and target domains can be modeled effectively as a residual function learned by an auxiliary network. Deep Residual Correction Networks (DRCN)~\cite{drcn-residual-connection} employ additional residual blocks that dynamically adjust feature representations by emphasizing the source domain classes most relevant to the target domain, facilitating improved partial domain adaptation.
Even beyond domain adaptation, the concept of learning corrective residual features has been utilized in tasks such as MRI reconstruction, where a deep convolutional neural network explicitly learns a correction feature to enhance reconstruction accuracy~\cite{error-correction-mri}. 

Inspired by these works, we propose applying residual correction to the problem of occluded gait recognition. Our proposed network dynamically integrates a learned residual correction feature with the holistic features, significantly improving performance on occluded gait sequences without compromising holistic recognition accuracy.





%% file: sec/sec3_method.tex
\section{Approach}

\begin{figure*}
    \centering
    \includegraphics[width=\linewidth]{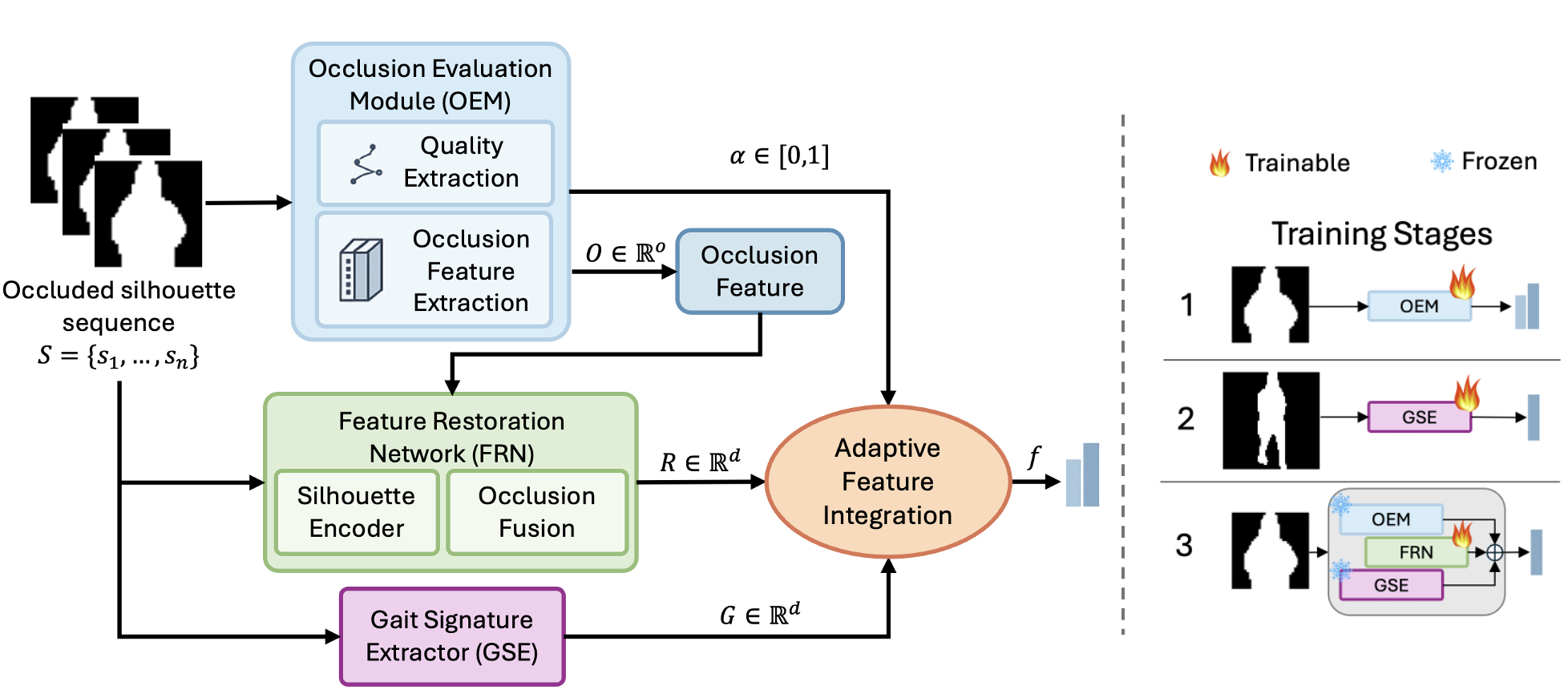}
    \caption{
    Our proposed RG-Gait framework. The left part shows the pipeline, and the right shows the multi-stage training setup. 
    GSE is any SOTA gait recognition backbone, which works well on holistic data but not on occlusions.
    The FRN, guided by OEM~\cite{occ_aware}, generates a residual deviation $R$ to adjust GSE features $G$ for occlusions.
    Adaptive integration of $R$ with $G$ ensures performance retention on non-occluded data.
    We train the framework over multiple stages.
    Stage 1 and stage 2 involve training the OEM and the GSE respectively.
    In stage 3, the FRN is trained to generate the residual deviation on the GSE features.
    }
    \label{fig:framework}
\end{figure*}

\subsection{Overview}
In this section, we present RG-Gait, a comprehensive framework for occluded gait recognition with holistic retention.
The core challenge in occluded gait recognition is the absence of complete information about a subject's walking pattern, which significantly hampers traditional methods.
Our method addresses this challenge through a three-module architecture that dynamically adapts to different levels of occlusion.

As illustrated in \cref{fig:framework}, \approach consists of three key components: (1) an Occlusion Evaluation Module (OEM) for detecting and quantifying occlusions, (2) a Feature Restoration Network (FRN) for compensating missing information, and (3) a Gait Signature Extractor (GSE) for capturing identity-specific gait features. These components work together to produce robust gait representations even when substantial portions of the silhouette are occluded.

\subsection{Problem Formulation}
Given as input a potentially occluded silhouette sequence $S = \{s_1, s_2, ..., s_n\}$ for a subject, where each $s_i \in \mathbb{R}^{H \times W}$ represents a binary silhouette frame and $n$ is the sequence length, our goal is to extract a discriminative feature vector $f \in \mathbb{R}^d
$ that remains effective for recognition regardless of occlusion conditions. 
If the input is RGB, it is converted to silhouettes using segmentation techniques.

The key challenge is that occlusions create inconsistencies in the silhouette sequence, corrupting the motion patterns essential for gait recognition. The missing information varies based on occlusion type (e.g., top, bottom, middle, or lateral occlusions) and levels.

\subsection{Gait Signature Extractor}
The GSE extracts identity-specific motion patterns from silhouette sequences. It processes the input silhouette sequence $S$ to produce a gait feature vector $G \in \mathbb{R}^{d}$; i.e.,
\begin{equation}
    G = \text{GSE}(S).
\end{equation}
Our method is model-agnostic, allowing the GSE to be implemented using any state-of-the-art gait recognition backbone. This flexibility enables our framework to benefit from advances in gait feature extraction while maintaining its core occlusion-handling capabilities.

\subsection{Occlusion Evaluation Module}
The OEM analyzes input silhouettes to detect occlusions and generates two critical outputs: an occlusion feature vector $O \in \mathbb{R}^{o}$ that encodes information about the occluded areas, and an occlusion quality score $\alpha \in [0,1]$ that quantifies the level of occlusion in the silhouette data.

The module consists of a convolutional backbone with two parallel branches for feature extraction and quality evaluation. Formally, for a silhouette sequence $S$, the OEM computes
\begin{equation}
    O, \alpha = \text{OEM}(S),
\end{equation}
where $O$ represents spatial information about occlusion patterns, and $\alpha$ is close to 1 for complete silhouettes and decreases with occlusion.

The OEM is trained with a combined objective
\begin{equation}
    \mathcal{L}_{\text{OAM}} = \lambda_1 \mathcal{L}_{\text{cls}} + \lambda_2 \mathcal{L}_{\text{reg}},
\end{equation}
where $\mathcal{L}_{\text{cls}}$ is a cross-entropy loss for occlusion type classification, and $\mathcal{L}_{\text{reg}}$ is an L2 loss for occlusion level estimation.

\subsection{Feature Restoration Network}
We model the occluded features to be a residual deviation from the holistic features generated by the GSE.
The FRN aims to generate this residual, to restore features that may be missing due to occlusions.
It takes both the original silhouette sequence $S$ and the occlusion feature $O$ as inputs and produces a restorative residual feature $R \in \mathbb{R}^{d}$ that compensates for this missing information.

The FRN is based on the same architecture as the GSE, that fuses silhouette and occlusion information through
\begin{equation}
    \label{eqn:FRN}
    R = \text{FRN}(S, O).
\end{equation}
The network consists of a silhouette encoder and a fusion module~\cite{occ_aware} that combines silhouette features with occlusion features, to output the residual correction feature vector $R$.

\subsection{Adaptive Feature Integration}
The final gait representation $f$ integrates information from both the GSE and the FRN, weighted by the occlusion quality score from the OEM; i.e.,
\begin{equation}
    f = G + \alpha \cdot R
\end{equation}
This adaptive integration allows the model to rely primarily on direct gait features when silhouettes are complete, while incorporating more restorative residual features when occlusions are significant. 
The quality score $\alpha$ serves as a confidence measure that dynamically balances the contribution of each feature type based on the detected occlusions.

\subsection{Training}
We train \approach over three stages.

\textbf{Stage 1.} Train the OEM using synthetic occlusions with known ground truth occlusion types and levels, like~\cite{mimicgait}.

\textbf{Stage 2.} Train the GSE on holistic silhouettes using standard triplet and cross-entropy losses, as done in~\cite{opengait, deepgaitv2, lin_gait_2021}.



\textbf{Stage 3.} Freeze the OEM and GSE weights and train the FRN using an end-to-end loss

\begin{equation}
    \mathcal{L}_{\text{FRN}} = \mathcal{L}_{\text{triplet}}(f) + \lambda_3 \mathcal{L}_{\text{xe}}(f)
\end{equation}

where $\mathcal{L}_{\text{triplet}}$ is a triplet margin loss, and $\mathcal{L}_{\text{xe}}(f)$ is a cross-entropy  subject classification loss after passing $f$ through a BNNeck layer~\cite{opengait}.


This multi-stage method ensures that each component learns its specialized function, and the final feature $f$ becomes an identifying gait feature regardless of the occlusion conditions or lack thereof.

\subsection{Inference}
During inference, the complete pipeline processes a potentially occluded silhouette sequence and outputs a robust feature vector $f$ for recognition. 
The model dynamically adjusts its confidence $\alpha$ on direct versus restorative features based on the detected occlusion, enabling effective recognition across various occlusion conditions without requiring explicit occlusion labels at test time.

\subsection{Implementation Details}
We train our models on 4x A5000 GPUs (GaitBase) or 8x A5000 (DeepGaitV2) with a batch size of (32,4).
We use the SGD optimizer with a weight decay of 0.0005 and an initial learning rate of 0.01, reducing the learning rate using the MultiStepLR scheduler. 
We use a margin of 0.2 in our triplet loss in both Stage 2 and Stage 3. 
We set $\lambda_1 = 0.1$ and $\lambda_2 = \lambda_3 = 1$ in our experiments.
We follow a similar strategy for occlusion-awareness as~\cite{occ_aware}, using the last \textit{FC} layer of the network for occlusion fusion.
Additional implementation details are included in the supplementary.

%% file: sec/sec4_experiments.tex
\section{Experiments}

\subsection{Datasets}

We experiment with two publicly available datasets GREW~\cite{grew} and Gait3D~\cite{gait3d}. We also evaluate our approach on the challenging BRIAR~\cite{briar} dataset. All of these datasets are collected in unconstrained environments, from multiple viewpoints and ranges. GREW and Gait3D are still publicly available to reproduce our results. 
BRIAR has been collected ethically with the informed consent of all the individuals involved in the data collection. 
We describe these datasets in more detail below.

\paragraph{Gait3D}
The Gait3D dataset is a large-scale publicly available benchmark designed for gait recognition with accompanying 3D representation data. 
It comprises 4,000 subjects and over 25,000 sequences captured from 39 cameras within an unconstrained indoor environment. 
The diverse camera placements and the indoor setting provide a comprehensive range of gait variations, including self-occlusions due to carrying objects, facilitating the development and evaluation of robust gait recognition algorithms \cite{gait3d}.

\vspace{-3mm}
\paragraph{GREW}
The Gait Recognition in the Wild (GREW) dataset is also a large-scale publicly available benchmark for gait recognition under real-world conditions. 
It contains 26,345 subjects, providing both silhouettes and human poses. 
The data was collected from thousands of hours of video streams across multiple cameras in open environments. 
GREW introduces various challenges such as clothing changes, carrying conditions, and diverse viewpoints for evaluating gait recognition systems in complex, real-world scenarios. 
Due to the unconstrained setting, the dataset also naturally contains occlusions, but there are no occlusion labels on the gait sequences  \cite{grew}. 

\begin{figure}
    \centering
    \includegraphics[width=\linewidth]{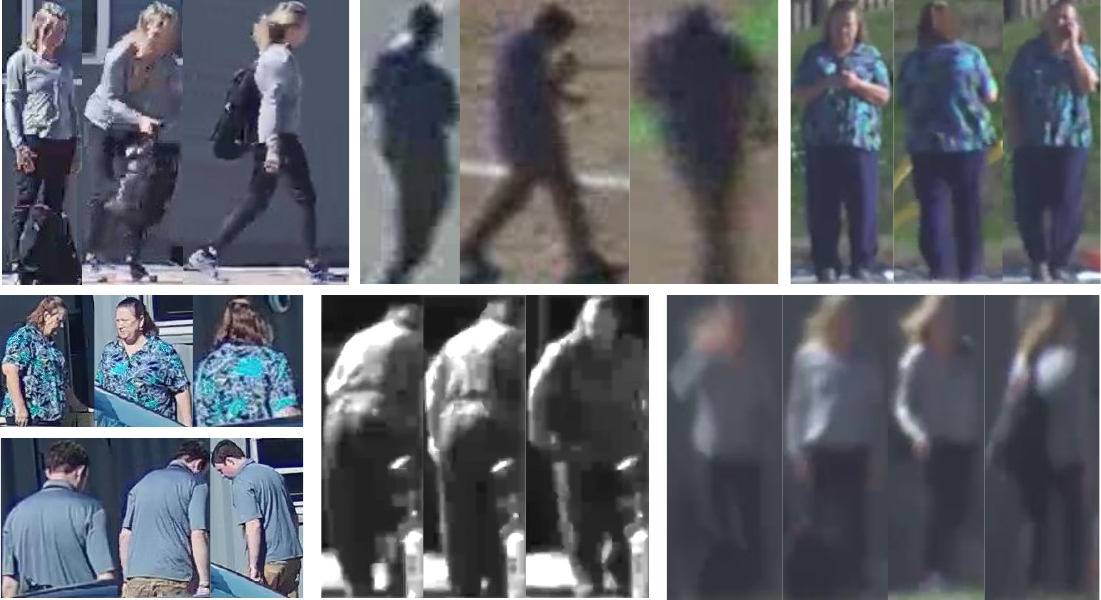}
    \caption{Sample frames from the BRIAR dataset illustrating the grayscale variations, physical turbulence, and occlusion challenges present across different distances and environments. Subjects have consented to the use of these images in publication.
    }
    \label{fig:visualization_briar}
\end{figure}

\vspace{-3mm}
\paragraph{BRIAR}
The BRIAR dataset \cite{briar} consists of images and videos captured under extreme conditions designed to challenge biometric recognition systems. It contains over 350,000 still images and 1,300+ hours of videos from more than 1,000 subjects, taken from ranges of up to 1,000m away. The dataset is partitioned into multiple subsets, each featuring different subjects, clothing variations, modalities, self occlusions due to heavy clothing and environmental challenges. 
BRIAR has three primary collection environments: indoor, outdoor, and aerial. Indoor data was captured in controlled settings with fixed camera positions and minimal interference, making it suitable for baseline training. Indoor activities include structured walking patterns along preset paths and random walking movements. The dataset also includes dedicated distractor sets containing subjects absent from both training and query sets, simulating real-world identification challenges. A key feature of BRIAR is its range diversity, with images captured at distances from close-range to 1,000 meters and face widths ranging from 10-32 pixels. Aerial footage was collected using various UAV platforms at altitudes up to 400 meters. The dataset introduces multiple challenges including occlusion (cones, poles, doors, boxes, backpacks), varying degrees of blur, physical turbulence effects that intensify with distance, and cross-clothing matching. \cref{fig:visualization_briar} showcases some sample RGB frames taken from the dataset.

\subsection{Synthetic occlusions}
\label{sec:synthetic-occ}

\begin{figure}
    \centering
    \includegraphics[width=0.8\linewidth]{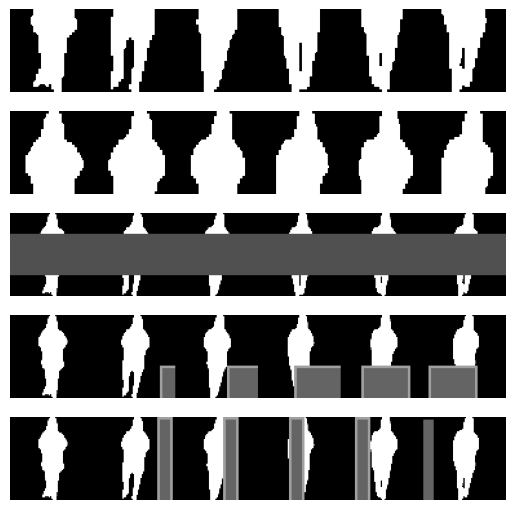}
    \caption{Synthetic occlusions we use in our work. The gray color of the patches is just shown for visualization, and the actual color of the occlusions is black in the binary silhouettes. We use top, bottom, middle and moving occlusions in our work.
    }
    \vspace{-3mm}
    \label{fig:synthetic-occ}
\end{figure}

Following previous works~\cite{mimicgait, occ_aware}, we experiment with synthetic occlusions of different types as shown in \cref{fig:synthetic-occ}.
Specifically, we randomly occlude the top, bottom, or middle part of the subject in the input silhouette. 
The size and position of the occlusion is chosen randomly during training and testing.  
We also introduce a black mask of varying size which can laterally move across the video at different speeds and occlude different parts of the subject within the same video.
We follow the same settings used in~\cite{mimicgait}, where up to 60\% of the subject may be occluded.
It is important to note that we introduce the occlusions randomly, without the need for paired (occluded, complete) data during training unlike some previous approaches~\cite{mimicgait}.

\input{tables/main_results}

\subsection{Metrics}
\paragraph{Rank-Retrieval} 
Following previous work~\cite{occ_aware, mimicgait}, we use the Rank-Retrieval accuracy metric to estimate gait recognition performance. 
Rank retrieval accuracy in gait recognition measures how well a probe sample (an unseen gait sequence) matches a corresponding identity in a gallery (a set of enrolled gait samples). 
It evaluates the rank at which the correct identity appears in the sorted list of gallery candidates, with Rank-1 accuracy indicating the percentage of times the correct match is the top-ranked result.
This task is relevant when multiple candidate matches need to be retrieved from the recognition system for later analysis.

\vspace{-3mm}
\paragraph{Verification}
Additionally, according to the BRIAR protocol~\cite{briar}, we also evaluate the models on the 1:1 verification task.
Verification in gait recognition determines whether a probe gait sample matches a specific gallery sample by computing a similarity score and verifying if it exceeds a predefined threshold. We choose this threshold by fixing the False Positive Rate to 0.01, and report the corresponding True Positive Rate as verification performance.
This task is useful when two different samples need to be verified if they belong to the same identity.

\vspace{-3mm}
\paragraph{Relative Performance (RP)}
Following previous works, we also compute the Relative Performance (RP) metric for the occlusion scenario. 
RP, proposed in \cite{mimicgait}, evaluates a gait recognition model's accuracy on occluded data relative to the upper bound performance of the original backbone on holistic data.
By normalizing occluded performance against holistic performance, RP offers a standardized measure of how effectively a method handles occlusions across different gait recognition backbones.

\vspace{-3mm}
\paragraph{Holistic Performance Retention (HPR)}
We evaluate the occluded recognition models directly on holistic data, and compare it to the holistic performance of the original backbone.
Ideally, both should be the same.
However, in \cref{fig:holistic-performance-retainment}, we observe this is not the case for most methods; occlusion training generally reduces holistic performance. 
Hence, we evaluate our method on holistic data to measure how much holistic performance is retained post occlusion training.

It is important to note the distinction between RP and HPR. 
For example, let us say we are given the backbone $B$, and a corresponding RG-Gait$_{B}$ built on $B$.
RP would compare the \textit{occluded} performance of RG-Gait$_{B}$ with the holistic performance of $B$.
However, HPR will compare the \textit{holistic} performance of RG-Gait$_{B}$ with the holistic performance of $B$ - to see if it is retained even after occlusion training.
Thus, RP is a measure of occluded performance, and HPR is a measure of holistic performance.


\subsection{Baselines}
We compare our method with baselines from previous works.
Baseline-1 refers to direct occluded evaluation of a model trained on holistic data, without any retraining.
Baseline-2 refers to simple training of these backbones on occluded data.
Occlusion Aware\cite{occ_aware} uses an auxiliary occlusion detector to insert occlusion awareness inside the gait recognition backbone.
MimicGait\cite{mimicgait} uses knowledge distillation techniques and an ideal teacher to guide the training of a student network on occluded data.

\input{tables/swingait}

\subsection{Results and Discussion}
Our main results on occlusion evaluations are summarized in \cref{tab:main}. We can see that our approach performs better than other works on the occluded gait recognition task, across different datasets and backbones. 
This shows that learning the correction feature as a residual can be effective for occluded gait recognition.

We also evaluate our method with the transformer-based SwinGait backbone \cite{deepgaitv2} in \cref{tab:swingait}. We compare it with the two baselines. Other works on occluded gait recognition do not report results on transformer backbones.

\begin{figure}
    \centering
    \includegraphics[width=\linewidth]{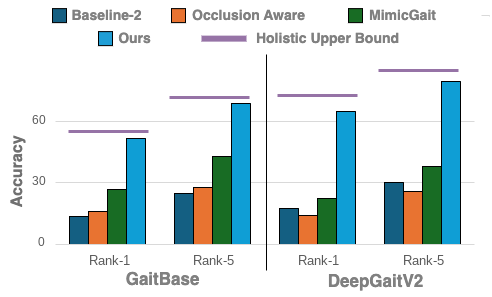}
    \caption{Holistic performance retention evaluation - i.e., taking models trained for occlusion and evaluating them on holistic data - on the GREW dataset.
    The upper bound is the performance of the original backbone on holistic data.
    While other methods drop significantly compared to the upper bound, our method is designed to retain holistic performance even after occlusion training.
    }
    \label{fig:holistic-performance-retainment}
    \vspace{-3mm}
\end{figure}

Our results on the holistic performance retention evaluation are shown in \cref{fig:holistic-performance-retainment}. In this evaluation, we evaluate the occlusion models on complete gait sequences. We observe that other occlusion methods drop significantly compared to the original holistic performance of the backbone.
Interestingly, we observe that in some cases, the occluded performance of previous methods is similar to their performance on holistic data - indicating that they are over-tuned for occlusions.
Regardless, our method is designed to retain most of the holistic performance of the original backbone. 


Another observation of note is that while our method is better than the baselines for all backbones, relatively shallow backbones like GaitBase perform better than deeper ones like DeepGaitV2 and SwinGait on the same dataset. 
This issue has also been observed by \cite{mimicgait}, and they mention that the sparse nature of the occluded silhouettes confuses the deep networks more compared to smaller networks. Our experimental results support this observation as well. 


\subsection{Ablations and Analysis}
In this section, we analyze the effect of the different components used in our framework - occlusion residual learning, adaptive feature integration, and occlusion awareness.

\input{tables/ablations}

\vspace{-2mm}
\paragraph{Occlusion Residual learning}
This component of our framework refers to the use of the FRN to learn the residual feature, subsequently used to adjust the features generated by the GSE.
Without the FRN and its occlusion training, no residual is learned and the method becomes equivalent to Baseline-1 - only the holistic network is used to directly generate features of the occluded inputs. 
The results are summarized in \cref{tab:ablations}. 
From the first two rows, we can observe that the FRN is essential for occluded performance.

\vspace{-3mm}
\paragraph{Adaptive Feature Integration}
The residual features generated by the FRN are regulated by the output from the OEM, $\alpha$, which is a measure of the amount of occlusion present in the input.
Without this regulation, the residual would always be considered in the gait signature, even when it might not be needed.
In \cref{tab:ablations}, we experiment with turning adaptive integration off. This is equivalent to fixing $\alpha = 1$, rather than choosing it based on the OEM. 
We observe that adaptive integration helps, and conclude that the residual learned by the FRN is more important for some scenarios and less important for others - and choosing $\alpha$ dynamically can help overall performance.

\vspace{-3mm}
\paragraph{Occlusion Awareness from OEM}
This component of our framework is inspired by \cite{occ_aware}. The OEM outputs a feature $O$ containing information about the occlusion patterns in the input. The FRN is guided by $O$ while generating the residual feature as described in \cref{eqn:FRN}.
In this experiment, we omit $O$ and let the FRN generate residual features without this guidance. 
From \cref{tab:ablations}, we observe that occlusion awareness, which is the guidance from the OEM, improves the generated residual feature.

\vspace{-3mm}
\paragraph{Combining all components}
From the last row of \cref{tab:ablations}, we observe that combining all the proposed components works best. Residual learning, adaptive integration, and occlusion awareness complement each other and achieve performance greater than their individual performances.

\subsection{Evaluating on new occlusions}

\begin{figure}
    \centering
    \includegraphics[width=\linewidth]{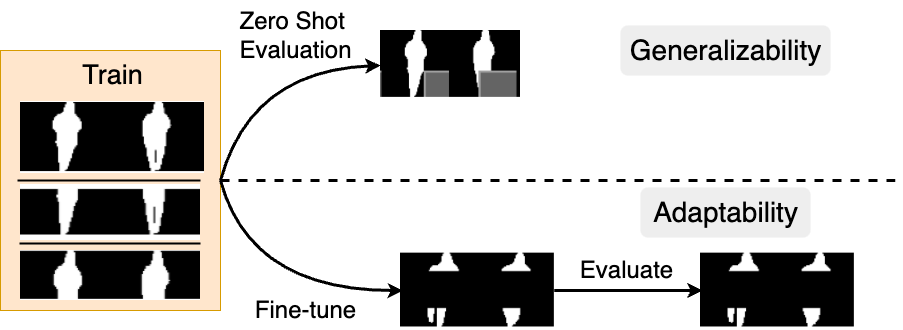}
    \caption{Illustration of Generalizability and Adaptability evaluations~\cite{mimicgait}, to test the flexibility of the approach on new occlusions.}
    \label{fig:generaize-adapt}
\end{figure}

We also conduct generalizability and adaptability tests on our model to evaluate whether it can work on new kinds of occlusions. These tests have been proposed in \cite{mimicgait} and evaluate the robustness of the method on different occlusion scenarios. 
For a gait recognition method to be deployed in a practical scenario, it must be able to handle a diverse set of occlusion types. However, it is not feasible to train a model on all possible occlusion types. 
Hence, these evaluations aim to quantify how well networks can work on new occlusion types. 
We describe these two types of evaluations, with an illustration of the evaluation setup shown in \cref{fig:generaize-adapt}.

\textbf{Generalizability} refers to how well a model generalizes to new occlusions it has not seen during training. 
Specifically, in this evaluation, we train a model for top and bottom occlusions and directly evaluate on middle and moving occlusions.
This evaluation is well suited for the scenario when the model has to perform on occlusions unseen during training, and evaluates the inherent ability of the algorithm to extend naturally to unseen occlusions.

\textbf{Adaptability} refers to how well an algorithm can perform on new occlusions after being given the opportunity to fine-tune on them.
This scenario evaluates how well the algorithm can extend its capability to a new occlusion type, if that particular type is anticipated to be more frequent in the environment the algorithm is deployed in.
Thus, in this evaluation, we fine-tune a model initially trained on top and bottom occlusions on the new occlusion types, and then evaluate on these new occlusions, like~\cite{mimicgait}.

We conduct these tests on dynamic and middle occlusions in \cref{tab:generalize-adapt}. \approach can generalize and adapt to newer occlusions better than other approaches.

\input{tables/generalizability_adaptability}

\subsection{Randomness in evaluation}

The evaluation method we use to evaluate our models on occluded data involves a certain degree of randomness, due to the random application of occlusions. As described in \cref{sec:synthetic-occ}, the occlusion type and the amount is chosen randomly for a video. 
As such, the input data is slightly different in every evaluation run.
Hence, we investigate whether this randomness significantly affects the performance across multiple evaluation runs. Similar to \cite{mimicgait}, we run 10 evaluations, each with different random seeds, and observe the mean accuracy and its standard deviation. 
Specifically, for the Gait3D dataset with the GaitBase backbone, we observe the distribution of Rank-1 accuracy to be $40.72 \pm 1.14$ in these 10 runs, indicating minor differences in each run.
Thus, we conclude that while the input data is different in each evaluation, its effect is generally averaged out across all sequences in the dataset.

%% file: tables/main_results.tex
\begin{table*}
\centering
\resizebox{\linewidth}{!}{%
\begin{tabular}{c|c|cc|cc|ccc}
\multirow{2}{*}{Backbone}   & \multirow{2}{*}{Method} & \multicolumn{2}{c|}{Gait3D}                                                            & \multicolumn{2}{c|}{GREW}                     & \multicolumn{3}{c}{BRIAR}                                              \\ 
\cline{3-9}
                            &                         & Rank-1                                    & Rank-5                                     & Rank-1                & Rank-5                & Rank-1                & Rank-20               & TAR@0.01 FAR           \\ 
\hline
\multirow{5}{*}{GaitBase \cite{opengait}}   & Baseline-1              & 7.6 (0.11)                                & 15.71 (0.19)                               & 14.85 (0.27)          & 25.55 (0.35)          & 1.34 (0.04)           & 12.05 (0.16)          & 2.46 (0.04)            \\
                            & Baseline-2              & 17.12 (0.24)                              & 31.43 (0.37)                               & 16.42 (0.30)          & 30.38 (0.42)          & 6.13 (0.16)           & 27.52 (0.36)          & 9.81 (0.17)            \\
                            & Occlusion Aware \cite{occ_aware}         & 18.22 (0.26)                              & 34.94 (0.41)                               & 22.55 (0.41)          & 37.95 (0.53)          & 8.70 (0.23)           & 35.38 (0.46)          & 12.83 (0.22)           \\
                            & MimicGait \cite{mimicgait}               & 22.72 (0.32)                              & 40.84 (0.48)                               & 28.38 (0.51)          & 45.43 (0.63)          & 10.93 (0.28)          & 42.25 (0.55)          & 12.92 (0.22)           \\
                            & \textbf{RG-Gait (ours)}        & \textbf{40.84 (0.58)}                     & \textbf{57.25 (0.68)}                      & \textbf{43.02 (0.78)} & \textbf{59.27 (0.82)} & \textbf{13.88 (0.36)} & \textbf{45.28 (0.59)} & \textbf{27.53 (0.47)}  \\ 
\hline
\multirow{5}{*}{DeepGaitV2 \cite{deepgaitv2}} & Baseline-1              & 2.2 (0.03)                                & 6.21 (0.07)                                & 3.72 (0.05)           & 6.85 (0.08)           & 2.38 (0.05)           & 11.4 (0.13)           & 1.8 (0.03)             \\
                            & Baseline-2              & 6.71 (0.09)                               & 14.21 (0.16)                               & 9.65 (0.13)           & 16.2 (0.19)           & 6.52 (0.13)           & 27.47 (0.32)          & 7.4 (0.11)             \\
                            & Occlusion Aware \cite{occ_aware}         & 14.01 (0.18)                              & 27.83 (0.32)                               & 13.38 (0.18)          & 22.87 (0.27)          & 7.39 (0.15)           & 32.74 (0.38)          & 7.12 (0.1)             \\
                            & MimicGait \cite{mimicgait}               & 16.82 (0.22)                              & 33.03 (0.38)                               & 14.38 (0.20)          & 24.93 (0.29)          & 11.49 (0.24)          & 44.71 (0.52)          & 20.73 (0.3)            \\
                            & \textbf{RG-Gait (ours)}        & \multicolumn{1}{l}{\textbf{41.94 (0.55)}} & \multicolumn{1}{l|}{\textbf{56.15 (0.64)}} & \textbf{35.6 (0.49)}  & \textbf{50.83 (0.6)}  & \textbf{29.86 (0.62)} & \textbf{47.06 (0.55)} & \textbf{43.38 (0.62)} 
\end{tabular}
}
\vspace{0.5mm}
\caption{\label{tab:main}
Comparison of RG-Gait with other baselines, with different backbones on different datasets. The values in (.) denote the RP metric proposed in \cite{mimicgait}. Our model performs better than other methods on occlusions, across different datasets and backbones. 
}
\end{table*}

%% file: tables/swingait.tex
\begin{table}
\centering
\resizebox{0.7\linewidth}{!}{%
\begin{tabular}{c|cc}
Method & Rank-1 & Rank-5 \\ 
\hline
Baseline-1 & 18.48 (0.25) & 21.9 (0.26) \\
Baseline-2 & 20.28 (0.28) & 27.75 (0.33) \\
\textbf{RG-Gait (ours)} & \textbf{26.2 (0.36)} & \textbf{40.15 (0.47)}
\end{tabular}
}
\vspace{0.5mm}
\caption{\label{tab:swingait}
Our results using the transformer based SwinGait \cite{deepgaitv2} backbone on the GREW dataset. 
We report the Rank-retrieval accuracy along with the Relative Performance (RP) metric in parentheses.
Our method performs better than the baselines on the occluded gait recognition.
}
\end{table}

%% file: tables/ablations.tex
\begin{table*}
\centering
\resizebox{0.6\linewidth}{!}{%
\begin{tabular}{c|ccc|cc}
\textbf{Method}  & \begin{tabular}[c]{@{}c@{}}\textbf{Residual }\\\textbf{learning}\end{tabular} & \begin{tabular}[c]{@{}c@{}}\textbf{Adaptive }\\\textbf{Integration}\end{tabular} & \begin{tabular}[c]{@{}c@{}}\textbf{Occlusion }\\\textbf{Awareness}\end{tabular} & \textbf{Rank-1} & \textbf{Rank-5}  \\ 
\hline
Baseline-1       & \ding{55}                                                                            & \ding{55}                                                                              & \ding{55}                                                                              & 7.6             & 15.71            \\
Vanilla residual & \checkmark                                                                           & \ding{55}                                                                              & \ding{55}                                                                              & 35.3            & 50.8             \\
Aware only       & \checkmark                                                                           & \ding{55}                                                                              & \checkmark                                                                             & 38.03           & 54.95            \\
Adaptive only   & \checkmark                                                                           & \checkmark                                                                             & \ding{55}                                                                              & 38.23           & 52.75            \\
\textbf{RG-Gait (ours)} & \textbf{\checkmark}                                                                  & \textbf{\checkmark}                                                                    & \textbf{\checkmark}                                                                    & \textbf{40.84}  & \textbf{57.25}  
\end{tabular}
}
\vspace{0.5mm}
\caption{\label{tab:ablations}
Ablation experiments on the different components of our framework. 
Residual learning refers to the use of the FRN, adaptive integration refers to regulation of the residual by the OEM, and occlusion awareness is the guidance of the OEM on the network.
These experiments are done on the Gait3D dataset.
Each of the components in our approach individually contributes to performance, and all the components complement each other to boost overall performance when used together. 
}
\end{table*}

%% file: tables/generalizability_adaptability.tex
\begin{table}
\centering
\resizebox{\linewidth}{!}{%
\begin{tabular}{c|c|cc|cc}
\multirow{2}{*}{Evaluation Type}             & \multirow{2}{*}{Method} & \multicolumn{2}{c|}{Middle}                              & \multicolumn{2}{c}{Dynamic}                        \\
                                  &                         & \multicolumn{1}{l}{Rank-1} & \multicolumn{1}{l|}{Rank-5} & \multicolumn{1}{l}{Rank-1} & \multicolumn{1}{l}{Rank-5}  \\ 
\hline
\multirow{3}{*}{Generalizability} & Occ Aware \cite{occ_aware}               & 17.93                      & 32.15                       & 21.27                & 36.5                        \\
                                  & MimicGait \cite{mimicgait}               & 21.73                      & 37.37                       & 26.77                & 42.9                        \\
                                  & \textbf{RG-Gait (ours)}                     & \textbf{37.7}              & \textbf{55.25}              & \textbf{39.12}       & \textbf{55.45}              \\ 
\hline
\multirow{3}{*}{Adaptability}     & Occ Aware \cite{occ_aware}               & 26.7                       & 43.82                       & 34.87                & 52.07                       \\
                                  & MimicGait \cite{mimicgait}               & 34.78                      & 52.75                       & 36.65                & 53.15                       \\
                                  & \textbf{RG-Gait (ours)}                    & \textbf{47.62}             & \textbf{63.57}              & \textbf{48.72}       & \textbf{64.92}             
\end{tabular}
}
\vspace{0.5mm}
\caption{\label{tab:generalize-adapt}
Results on the generalizability and adaptability tests. Generalizability refers to zero shot evaluation on new occlusion types, and adaptability refers to fine-tuning on previously unseen occlusion types.
We experiment with middle and dynamic occlusion types, and observe that our approach performs better than previous works in these scenarios.
}
\vspace{-2.5mm}
\end{table}

%% file: sec/sec5_conclusion.tex
\section{Limitations and Future Work}


While our method significantly improves occluded performance, it does not implement perfect holistic retention. There is still a gap between the holistic upper bound of the models and our approach. Perhaps this can be solved by using an ensemble of different networks, or a more complex feature fusion than our adaptive feature integration strategy. We leave this investigation for future work.

Further, while we have a diverse set of synthetic occlusions, these occlusions are still different from real-world occlusions. We were unable to evaluate our model on real occlusions because of the lack of a large-scale occlusion-focused dataset for gait recognition, which can also be a potential direction for future work.

\section{Conclusion}
In this work, we proposed RG-Gait, a model-agnostic framework for residual correction in occluded gait recognition.
Current works on occluded gait recognition are not able to retain holistic performance. 
We model the occluded features to be a residual deviation from the holistic features, and design \approach to learn this residual and adaptively integrate it with the holistic features.
Through extensive evaluation on outdoor datasets, we showed that our approach outperforms other works on occluded gait recognition while retaining holistic performance.

%% file: sec/sec6_acknowledgement.tex
\vspace{-3mm}
\paragraph{Acknowledgements }
This research is based upon work supported in part by the Office of the Director of National Intelligence (ODNI),
Intelligence Advanced Research Projects Activity (IARPA), via [2022-21102100005]. The views and conclusions contained herein are those of the authors and should not be interpreted as necessarily representing the official policies, either expressed or implied, of ODNI, IARPA, or the U. S. Government. The US. Government is authorized to reproduce and distribute reprints for governmental purposes notwithstanding any copyright annotation therein.

%% file: sec/supp.tex
\clearpage
\setcounter{page}{1}
\setcounter{section}{0}

\maketitlesupplementary

\section{Introduction}
In this supplementary material, we provide additional visualizations of the BRIAR dataset. 
Next, we provide more implementation details about the preprocessing techniques we employed and details required to reproduce the training of RG-Gait. 
We also elaborate on the architectural details of the OEM, and provide additional clarification about the evaluation protocol used for the GREW dataset.

\section{BRIAR examples}
\begin{figure*}[t]
    \centering
    \includegraphics[width=\linewidth]{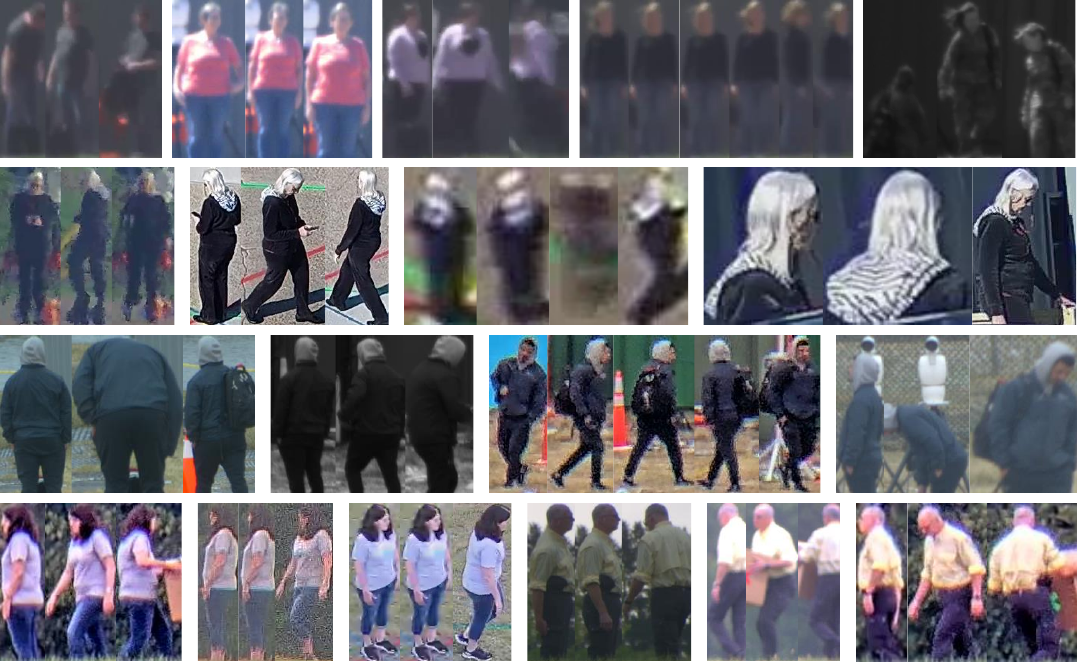}
    \caption{Visualization of challenging conditions in the BRIAR dataset. The images showcase how increasing distance (from left to right) affects recognition difficulty due to natural grayscale variations, physical turbulence, and reduced resolution. The 2nd row demonstrates examples of different imaging conditions including close-range captures, medium-range outdoor captures, and long-range captures (500-1000m). The 3rd and 4th rows illustrate various occlusion challenges (cones, poles, doors, backpacks) and blur effects present in the dataset. These diverse challenges collectively simulate real-world scenarios for robust evaluation of person re-identification algorithms.}
    \label{fig:SM-BRIAR}
\end{figure*}

\cref{fig:SM-BRIAR} illustrates the diverse challenges present in the BRIAR dataset across varying distances and conditions. As shown in the visualization, BRIAR exhibits natural grayscale variations and physical turbulence effects that become more pronounced as distance increases. Images captured at extreme ranges (1,000m) demonstrate significant degradation, with subjects sometimes blending into the background. \cref{fig:SM-BRIAR} also highlights various occlusion challenges incorporated in the dataset, including cones, poles, doors, boxes, and backpacks. These occlusions, either appearing individually or in combination, realistically reflect the complexity of unconstrained person re-identification scenarios. Additionally, the visualization demonstrates how environmental factors and camera positions contribute to blur and distortion effects. These examples underscore why BRIAR serves as a particularly rigorous benchmark for evaluating biometric recognition algorithms under challenging real-world conditions.

\section{Implementation Details}

\paragraph{Preprocessing:}
For an input video $S^i$ in RGB modality, we first generate binary silhouette masks using Detectron2 \cite{wu2019detectron2}, isolating the subject within each frame. This masking step is essential to remove background, color, and texture variations, which could otherwise adversely affect gait recognition performance. 

After extracting these masks, we center the detected subject and uniformly resize each frame to dimensions $H \times W$, following the preprocessing approach described in \cite{opengait}. For frames in which no subject is detected, we insert an empty black frame to ensure temporal consistency across the video sequence.

\paragraph{RG-Gait training: }


We utilize the \textit{fixed\_ordered} strategy for frame sampling, extracting a total of $n=30$ frames from each gait sequence~\cite{opengait}. For training GaitBase and DeepGait, we adopt a batch size configuration of $(32,4)$, where $(p,k)$ denotes $p$ distinct subjects with $k$ sequences each. Due to GPU memory constraints, we adjust the batch size for SwinGait to $(20,4)$. The total number of training iterations aligns with those used in \cite{opengait}, specifically $180,000$ for GREW and $60,000$ for Gait3D, while extending it to $300,000$ iterations for BRIAR. Additionally, following \cite{deepgaitv2}, SwinGait is initialized with pretrained weights from the earlier convolutional layers of DeepGaitV2.

\section{Occlusion Evaluation Module (OEM)}
OEM is a convolutional neural network comprising three convolutional layers followed by one hidden linear layer, and two parallel linear heads dedicated to classification and regression tasks, respectively~\cite{mimicgait}. The detailed architecture of OEM is outlined in \cref{tab:oem-arch}.

\input{tables/oem-arch}

Depending upon the variety of occlusion types targeted during training, the classification head categorizes inputs into either specific occlusion classes or a no-occlusion class. OEM is trained with cross-entropy loss through the classification head, enabling the network to become aware of different occlusion categories, and in the process, learns occlusion-relevant features.

These categories represent broad classes of occlusions, within which the intensity of synthetic occlusion can vary. The regression head, on the other hand, produces outputs designed to approximate $0$ in cases of no occlusion, and a value $x$ corresponding to an occlusion magnitude of $x$.

\section{GREW evaluation protocol}
The GREW dataset~\cite{grew} lacks identity annotations within its probe set, preventing direct local evaluation of models. According to the standard evaluation practice, prediction scores for gallery-probe video pairs must be submitted to the GREW official website, where accuracy is computed externally. This approach poses challenges for our experiments, particularly due to synthetic occlusions that introduce variations distinct from the original data.

To address this limitation, we adopt an alternative evaluation scheme for GREW, originally proposed by~\cite{opengait} and subsequently employed in recent studies on occluded gait recognition~\cite{occ_aware, mimicgait}. This revised protocol facilitates local evaluation by leveraging the fact that the GREW test set comprises 6,000 subjects, each with exactly two sequences. One sequence per subject is designated as gallery data, while the other serves as probe data, resulting in a balanced gallery-probe configuration of 6,000 videos each. The evaluation is then performed as a rank-based retrieval task, allowing calculation of standard Rank-K performance metrics. The unlabeled probe videos originally provided by GREW are disregarded in this adapted evaluation strategy.



%% file: tables/oem-arch.tex
\begin{table}[h]
\centering
\resizebox{\linewidth}{!}{%
\begin{tabular}{c|c|c}
\textbf{Layer Name} & \textbf{Input shape} & \textbf{Output Shape}  \\ \hline
Conv1               & 64 $\times$ 64 $\times$ 1          & 64 $\times$ 64 $\times$ 32           \\
ReLU, MaxPool1      & 64 $\times$ 64 $\times$ 32         & 32 $\times$ 32 $\times$ 32           \\
Conv2               & 32 $\times$ 32 $\times$ 32         & 32 $\times$ 32 $\times$ 64           \\
ReLU, MaxPool2      & 32 $\times$ 32 $\times$ 64         & 16 $\times$ 16 $\times$ 64           \\
Conv3               & 16 $\times$ 16 $\times$ 64         & 16 $\times$ 16 $\times$ 128          \\
ReLU, MaxPool3      & 16 $\times$ 16 $\times$ 128        & 8 $\times$ 8 $\times$ 128            \\
AdaptiveAvgPool     & 8 $\times$ 8 $\times$ 128          & 128                    \\ \hline
FC1                 & 128                  & 64                     \\
Classification Head                & 64                   & 6      \\
Regression Head     & 64            & 1
\end{tabular}
}
\vspace{1mm}
\caption{\label{tab:oem-arch} The architecture of the OEM. It is a three layer convolutional neural network, with one hidden linear layer, and two parallel linear heads which can predict the type and amount of occlusion in the input, corresponding to occlusion feature extraction and input quality extraction.}
\end{table}